%% file: root.tex
\documentclass[letterpaper, 10 pt, conference]{ieeeconf}
\IEEEoverridecommandlockouts
\usepackage{graphicx} 
\usepackage{amsmath} 
\usepackage{amssymb}  
\usepackage{multirow}
\usepackage{booktabs}
\usepackage{xcolor}
\usepackage{algorithm}
\usepackage[noend]{algpseudocode}
\usepackage{pifont}
\usepackage{listings}
\usepackage{hyperref}
\hypersetup{
  colorlinks,
  allcolors=.,
  urlcolor=black,
}

\newcommand\blfootnote[1]{%
  \begingroup
  \renewcommand\thefootnote{}\footnote{#1}%
  \addtocounter{footnote}{-1}%
  \endgroup
}

\title{\LARGE \bf Assembly Planning from Observations\\ under Physical Constraints}

\author{Thomas Chabal$^{1, 2}$, Robin Strudel$^{1, 2}$, Etienne Arlaud$^{1}$, Jean Ponce$^{1, 2, 3}$, Cordelia Schmid$^{1, 2}$
\thanks{${1}$ INRIA, ${2}$ Département d’informatique de l’ENS, École normale supérieure, CNRS, PSL Research University.
${3}$ Center for Data Science, New York University. Corresponding author: thomas.chabal@inria.fr}
}

\newcommand{\cmark}{\ding{51}}
\newcommand{\xmark}{\ding{55}}

\begin{document}
\def\baselinestretch{0.994}\selectfont

\maketitle
\thispagestyle{empty}
\pagestyle{empty}

\begin{abstract}
This paper addresses the problem of copying an unknown assembly of primitives
with known shape and appearance using information extracted from a single
photograph by an off-the-shelf procedure for object detection and pose
estimation. The proposed  algorithm uses a simple combination of physical
stability constraints, convex optimization and Monte Carlo tree search to plan
assemblies as sequences of pick-and-place operations represented by STRIPS
operators. It is efficient and, most importantly, robust to the errors in object
detection and pose estimation unavoidable in any real robotic system. The
proposed approach is demonstrated with thorough experiments on a UR5
manipulator.
\blfootnote{Project website: \url{https://www.di.ens.fr/willow/research/assembly-planning/}}
\end{abstract}


\section{Introduction}
The problem of replicating with a robot an assembly of known shapes
using photos as guidance is as old as AI itself, and its first
solution probably dates back to the famous ``copy demo'' of Patrick
Winston and his MIT AI Lab colleagues in 1970 \cite{winston1970}.  This
is the problem once again addressed in this presentation, but this
time the focus is on the difficulties associated with the
(unavoidable) failures of the vision system in planning the assembly
of the copy, and how to tackle them.

Concretely, we assume that we are given a set of {\em primitives} in
the form of 3D shapes with known shape and appearance, and two color
pictures, a {\em target} photo $I$ of some stable stack of these objects on a table in the
workspace of the robot, to be copied, and a {\em layout} image
$J$ of available
primitives laid elsewhere on the table. Some off-the-shelf vision module
extracts from the photos the identity of the primitives as well as
their position and orientation in space, and passes them to the
planner which then constructs a sequence of pick-and-place operations
represented by STRIPS operators, which is then executed by the robot
to copy the initial stack, see Figure~\ref{fig:teaser}.

We use {\em CosyPose}~\cite{labbe2020b} as our vision module. As of this
writing, it represents the state of the art in object detection and
pose estimation and has won five awards at the ECCV'20 6D object pose
challenge.  Yet, as any vision system, it is prone to errors such as
missed detections and inaccurate pose estimation when deployed in real
(if simple) robotic tasks such as addressed in this paper, notably due
to the numerous occlusions present in typical assemblies (some of
the primitives involved may in fact be completely hidden from the
camera).
Though we rely here on a single target photo, the use of multiple views may help reconstruction, but vision failures due to occlusions would still remain. Our work is therefore not specific to single observation settings.

We use a simple combination of physical stability constraints, convex
optimization and Monte Carlo tree search to build pick-and-place plans
robust to such failures of the vision system, see Figure~\ref{fig:overview}. Our assembly planner has
been implemented on a UR5 robot manipulator, and our experiments
demonstrate that it can cope with cases where multiple primitives in the target assembly go undetected (see Figures~\ref{fig:structure_poses_for_eval} and
\ref{fig:other_structures}).

\input{figures/teaser}
\input{figures/overview}


\section{Related Work}

\subsection{Assembly}

Building structures with a robot is a longstanding problem in robotics \cite{winston1970}. 
Previous art tackles assemblies composed of either a single \cite{ikeuchi1994, stevsic2020} or multiple steps \cite{funk2021, lin2021, pashevich2020}, starting from different specifications of the goal. Existing methods based on observations perform single step assemblies. Ikeuchi et al. \cite{ikeuchi1994}
take as input images before and after a change and compare them to determine a single action leading to the change, then execute it on a robot. Stevsic et al. \cite{stevsic2020} learn to predict the 6D pose of the known object to place next in a given structure.
Another line of work focuses on multi-steps assemblies with more expressive target representations. \cite{funk2021} and \cite{lin2021} describe scenes as a graph of contacts, where objects are nodes and edges are contacts. They learn a graph neural network outputting either pick and place positions for known objects and target poses \cite{lin2021} or Q-values to search for an assembly sequence and stack cubes to build a target mesh \cite{funk2021}.
Similarly, \cite{pashevich2020} propose to disassemble predefined goal structures known through their CAD models, the disassembly then provides a sequence of actions that is inverted to obtain a construction plan used to learn RL policies.
In this work, we rely on a single goal image to plan complex multi-steps assemblies. We have access to the CAD models of primitives but infer their identities and poses only from the input image. Our method is more widely applicable and not specific to one structure. The problem is more challenging as an image only provides partial, incomplete information about the goal structure due to unavoidable mistakes in the identification and pose estimation, for example due to occlusions or failures from the pose estimation system. 
To deal with these issues, we propose a novel approach exploiting physical constraints to recover missing objects and their poses from a set of available primitives and get a feasible assembly sequence.

\subsection{Pose Estimation}

Object pose estimation is an active field in computer vision.
Existing approaches base their estimation on either  keypoints~\cite{he2020,rothganger2006},  templates~\cite{wang2019,xiang2018} or render-and-compare~\cite{li2018, rad17}.
We use CosyPose~\cite{labbe2020b}, a render-and-compare estimator, to extract poses from the input images, see for example Fig.~\ref{fig:structure_poses_for_eval}.
It is a two-stage method that first detects object bounding boxes with a Mask-RCNN backbone \cite{he2017} and then regresses the pose of each object by comparing synthetically rendered images to the target object.

\subsection{Task Planning}

Multi-steps assembly requires to plan a sequence of actions, a problem typically addressed with task planning for which high-level tools were developed in the last decades.
STRIPS \cite{fikes1971} introduces predicates defining object states and operators to modify these states.
STRIPS operators have then been extended by \cite{garrett2020, pddl, Pednault1987FORMULATINGMD} to encompass a richer set of states and interactions.
Task planning has also been combined with motion planning in task and motion planning (TAMP), a more general framework to perform both high-level and low-level reasoning on motions \cite{lozano2014, toussaint2015}.
Yet, these methods usually assume full knowledge of both the environment the robot acts in and the goal to reach, which limits their applicability to controlled and predefined scenes.
Several approaches acquire this knowledge by relying on depth maps on simple scenes without occlusions \cite{driess2020b, driess2020a} or on QR codes placed on objects to recover their poses \cite{migimatsu2019}. We propose to build on STRIPS planning tools and develop a method that is able to cope with partial and imperfect knowledge of the target which is acquired only from a photograph.

\subsection{Monte-Carlo tree search (MCTS)}

Monte-Carlo tree search (MCTS) has been used with great success in notoriously hard boardgames such as Go~\cite{silver2016} and chess~\cite{silver2017} and, more recently, in robotic planning tasks~\cite{labbe2020a,paxton17,toussaint17}. We have chosen it as the backbone of our planning algorithm rather than other classical uninformed or informed tree-search algorithms (as used in classical STRIPS planners for example) since it only relies on a reward function without handcrafted heuristics or specific domain knowledge  to effectively explore the state space by learning a value function.


\section{Our approach}
\label{sec:method}
We assume throughout that the assemblies we want to copy are formed by stacking objects chosen from a set $\mathcal{B}$ of $N$ primitive shapes. Our plans are {\em linear}, in the sense that objects are placed on top of one another in sequence, so each plan consists of exactly $N$ pairs $(p_i,q_i)$, where $1\le i\le N$, $p_i$ denotes the {\em pick pose} of primitive number $i$ in the layout image $J$ where the object lies on the table,
and $q_i$ is the {\em place pose} of that primitive in the final
assembly, estimated by either the vision module or our planner
(Fig.~\ref{fig:overview}).

We assume that the primitives are well separated in the layout image, so that our vision module identifies them correctly and assigns them accurate pick poses $p_i$ (this has been the case in our experiments).
As the objects we consider are roughly parallelepipeds, we grasp them vertically from the top along their longest side.
On the other hand, we denote by $\mathcal{V}$ the {\em visible} primitives in the target image $I$, that is, those that have been identified and assigned a place pose $q_i$ by the vision module, and by $\mathcal{H}=\mathcal{B}\setminus\mathcal{V}$ the
remaining {\em hidden} primitives. The place poses $q_i$ associated with hidden primitives are recovered by our planner using physical
stability constraints.\footnote{In practice we of course apply some global offset to place poses to avoid collisions with the original structure.}

We formulate an assembly plan as a sequence of STRIPS operators
guided by a MCTS search as presented in Section \ref{subsec:task_planning} (Fig.~\ref{fig:overview}b). 
The STRIPS based plan provides a candidate arrangement between objects and we still need to obtain precise poses
for each object. To do so, we optimize over the object poses constrained by the candidate arrangement
and physical stability as described in Section~\ref{subsec:reward} (Fig.~\ref{fig:overview}c).
Lastly, a reward is computed according to the match between the candidate and target structure (Fig.~\ref{fig:overview}d).

Note that if all object poses could be detected accurately, we could just stack the objects according to their pose in the structure in ascending height order without searching for an assembly plan. The difficulty of our problem comes from the unavoidable failures of the vision system.

\subsection{From target image to object poses}
\label{subsec:pose_estimation}

As mentioned before we use a vision module to identify (some of)
the objects forming the target assembly and estimate their place
pose $q^{\mathcal{V}} = (q_{j})_{j\in\mathcal{V}}$. We use
CosyPose~\cite{labbe2020b} for that purpose in our experiments. Although it achieves state-of-the-art performance on the recent challenge BOP \cite{hodan2018},
its results are of course imperfect in real-world conditions, see Fig. \ref{fig:structure_poses_for_eval}, \ref{fig:other_structures}: it may miss some objects, hallucinate others, or estimate incorrect poses. 
We filter out objects whose detection scores are below a high confidence threshold. We obtain partial information about the structure to assemble in 3D from $q^{\mathcal{V}}$ and we show in the next sections how to build a plan and recover the poses $q^{\mathcal{H}}$ of objects missed by the object detector.

\subsection{Assembly planning with STRIPS based MCTS}
\label{subsec:task_planning}
Single-view observations of assemblies often include objects being partially or even
completely occluded. We exploit the information on the pose of the detected objects $q^{\mathcal{V}}$ to define a reward function, which increases with the
number of objects correctly placed in the known poses.  We then find a sequence
of STRIPS operators assembling the complete set of objects into a physically
feasible structure which maximizes the reward function and allows to recover the pose of hidden objects $q^{\mathcal{H}}$.

\noindent\textbf{STRIPS:}
MCTS is a search method over a discrete set of actions. 
The definition of this set impacts the efficiency of the algorithm.
For instance, discretizing the search
in a 3D volume results in a very large action space. 
We instead consider STRIPS operators
\cite{fikes1971} as the set of admissible actions, a pragmatic choice that
allows to reason about object relations in 3D while limiting the search space. Here, we restrict the target objects to be placed along the $X$
or the $Y$ axis in a 3D Manhattan world, with rotations being multiples of
$\frac{\pi}{2}$. This hypothesis could be avoided by defining more general STRIPS operators.

In more details, STRIPS represents the state of objects  by a set of predicates that define high-level geometric
relations among them as depicted in
Figure \ref{fig:overview}b. 
Given objects
$(a, b, c)$ we consider the following states: \textit{OnTable(a)}, $a$
is on the table; \textit{Clear(b)}, no object above $b$; $On(a, b)$,
$a$ is on $b$; \textit{OnAlongX(a, b, c)} (resp.
\textit{OnAlongY(a, b, c)}), $a$ is on $b$ and $c$ along the $X$ axis (resp. $Y$
axis); and \textit{Rot(a)}, $a$ is rotated by 90°. The state of objects can be
modified by $M = 4$ operators applied only on objects that have not been
moved yet: \textit{PutOn(a, b)} put $a$ on $b$; \textit{PutOnAlongX(a, b,
c)} (resp. \textit{PutOnAlongY(a, b, c)}) put $a$ on $b$ and $c$
along the $X$ axis (resp. $Y$ axis); \textit{Rotate(a)} rotate $a$ by 90°.
With this representation, there are at most $MN$ possible actions
at the root node, a number that shrinks after a few actions and enables an
efficient search.

\noindent\textbf{MCTS:} We perform a guided search over sequences of STRIPS
operators with MCTS as shown in Figure \ref{fig:overview}b.
A node of the tree
corresponds to a STRIPS state and a branch is an admissible STRIPS operator. We
denote by $n(s)$ the number of visits of node $s$, $g(s, a)$ the successor of
$s$ given action $a$ and $R(s)$ the cumulative reward or return of a node.  To
guide the search and select actions, we use the standard upper confidence bounds applied to trees (UCT) formula:

\begin{align}
    U(s, a) = Q(s, a) + C\sqrt{\frac{\log n(g(s, a))}{n(s)}},
    \label{eq:uct_formula}
\end{align}
where $Q(s, a)= \frac{R(g(s, a))}{n(g(s, a))}$ is the empirical estimate of the
return and $C$ is the exploitation-exploration trade-off constant, which we set
to $\sqrt{2}$. UCT controls the trade-off between exploration of unvisited nodes and exploitation of 
nodes with highest returns. Once we reach a leaf node where all actions have not been
explored yet, we execute random operators until all the target objects have been moved.
For each node, we filter operators to exclude physically unfeasible structures or structures that
do not match the structure from $\mathcal{V}$.
The sequence of STRIPS operators define the topology of the candidate structure $\mathcal{S}$, e.g.
the objects forming the base, the ones on top of them or the objects linking
two stacks of an arch. We still need to obtain precise poses of objects to define a
reward for the candidate structure. To do so we solve an optimization problem on the set of object poses with constraints defined from the sequence of STRIPS operators described in the next section and illustrated in Figure \ref{fig:overview}c.
Once the reward is computed, it is backpropagated from the leaf node to update information 
in the nodes of the sequence.

\subsection{Estimating missing poses by leveraging physical constraints}
\label{subsec:reward}

\noindent\textbf{Optimizing into an assembly plan:}
Given the set of visible poses $q^{\mathcal{V}}$ and a candidate sequence of STRIPS operators $\mathcal{S}$, the goal
is to find a precise assembly plan where all candidate object poses $q^{\mathcal{S}}$ are defined. 
The optimization objective is the sum of errors between $q^{\mathcal{S}}$ and the set of visible object poses $q^{\mathcal{V}}$ extracted from the
target image $I$. We define the problem
$(\mathcal{P})$ as follows:

\begin{equation}
    \min_{q^{S}_1, ..., q^{S}_N \in \mathbb{R}^6} \sum_{i\in\mathcal{V}} \|q^{\mathcal{S}}_i - q^{\mathcal{V}}_i\|_2^2 \quad
    s.t. \quad  \mathcal{C}(q^{\mathcal{S}}_1, ..., q^{\mathcal{S}}_N) \leq 0,
    \label{eq:reward_problem}
\end{equation}

where $\mathcal{C}(q^{S}_1, ..., q^{S}_N)$ corresponds to physical constraints
described in the next paragraph. Note that this is a convex optimization problem
as the criterion is convex and the constraints are linear. It has at least one
solution unless the constraints $\mathcal{C}(q^{S}_1, ..., q^{S}_N)$ cannot be
satisfied, in which case the stack is unfeasible.  However, in most cases where
the target is partially known, i.e.  $\mathcal{H} \neq \emptyset$, the objective is not strictly
convex and has an infinite number of solutions. Constraining the problem with
$\mathcal{C}$ restricts the optimization to a set of physically feasible
solutions, a highly desirable property when searching for an assembly plan.
$(q^{S}_1, ..., q^{S}_N)$ then corresponds to a candidate set of poses to
assemble the target according to the STRIPS sequence from Section \ref{subsec:task_planning}. 

\noindent\textbf{Stability and penetration constraints:} We define a set of stability
constraints for each of the STRIPS predicates defined in Section \ref{subsec:task_planning}.
Each constraint is defined thanks to linear inequalities, for example when
stacking a block $j$ on top of other blocks $i \leq j$ with the $\textit{PutOn}$ operator, we
use the following constraints. Let $(x_{i}, y_{i}, z_{i})$ be the center of mass of
object $i$ and $(s_{i,x}, s_{i,y}, s_{i,z})$ the size of its bounding box, the constraints
can be written as:
\begin{equation}
\left\{
\begin{array}{l}
    x_i - s_{i,x}/2 \leq x_{j} \leq x_i + s_{i,x}/2, \\
    y_i - s_{i,y}/2 \leq y_{j} \leq y_i + s_{i,y}/2, \\
    z_{i+1} = z_i + (s_{i,z} + s_{i+1,z})/2,
\end{array}\right.
\end{equation}
for all $i = 1,...,j-1$. These constraints enforce stability of the 
structure built by ensuring that the center of mass of the stacked object $j$ is placed at a
stable position. We denote $\mathcal{C}(q_1, ..., q_j) \leq 0$ the set of stability
constraints to be satisfied by the poses $q_{1}, ..., q_{j}$. Similar constraints are written for each operator,
each ensuring that objects are assembled in a stable structure. These constraints are specific to parallelepipeds, and assembling complicated shapes would require defining more general stability conditions, possibly combined with the use of a non-linear solver.

We also add object penetration related constraints to ensure physically plausible solutions. For instance,
if objects $a$ and $b$ with enclosing bounding boxes of size $(s_a, s_b)$ are interpenetrating
and $x_a \leq x_b$, we may assume that $a$ must be on the left of $b$ and add
the new constraint $x_a + s_a/2 \leq x_b - s_b/2$ to
$\mathcal{C}(q^{\mathcal{S}}_1, ..., q^{\mathcal{S}}_N)$.  In this way, we linearize
penetration-avoidance constraints and can solve the previous problem with our
updated constraints in order to recover penetration-free solutions.

\noindent\textbf{Reward:} We consider an assembly to be successful when all
the poses of visible objects $\mathcal{V}$ are matched, i.e. when
$\|q^{\mathcal{S}}_i - q^{\mathcal{V}}_i\|_2 \leq \epsilon$ 
for all $i \in \mathcal{V}$, see Figure~\ref{fig:overview}d.

We define the reward as the fraction of objects matched to target poses as follows:
\begin{equation}
  r(q^{\mathcal{S}}_1, ..., q^{\mathcal{S}}_N) = \frac{1}{|\mathcal{V}|} \sum_{i\in\mathcal{V}} 1_{\|q^{\mathcal{S}}_i - q^{\mathcal{V}}_i\|_2 \leq \epsilon},
  \label{eq:reward_formula}
\end{equation}
where 1 corresponds to a candidate structure matching all visible objects.


\section{Experiments}
\input{figures/dataset}
\input{figures/structures_eval}
\input{tables/constructions}
\input{tables/collision_removal}
\input{tables/mcts_random_search}
\input{tables/real_robot}

\subsection{Experimental setup}
\label{subsec:setup}
We build structures made of objects selected from the T-LESS dataset \cite{hodan2017}, as shown on Figure \ref{fig:dataset}.
We record RGB images with two RealSense D435 cameras pointing respectively to the target structure and the primitives laid on the table (Fig. \ref{fig:overview}). As mentioned earlier, we use CosyPose~\cite{labbe2020b} as our vision module, with a 
95\% threshold on detection confidence in our experiments. The poses of both cameras are assumed to be known through calibration in the robot
coordinate system. Our code is written in Cython \cite{behnel2011} and relies on CVXPY \cite{agrawal2018, diamond2016} to solve equation (\ref{eq:reward_problem}).

The assemblies are performed by a UR5 robot manipulator with a parallel gripper that successively picks primitives on the table at positions estimated by CosyPose and places them at the poses computed by our planner. We perform both off-line experiments,
using a real image of a scene as input but without an actual run on the robot, and on-line ones with the robot executing
the plan. The robot trajectory itself is computed by an off-the-shelf planner~\cite{kuffner2000}. Our off-line experiments use one target image per structure and average the metrics on 20 random seeds for the search. 
For experiments on the robot, we build between 10 and 15 instances of each structure where the pose of each object is slightly randomized manually to test the method robustness. For a given structure, we shoot all the target photos from the same viewpoint, which we select to observe a large number of objects. Changing the camera position could result in more occlusions and lower-quality pose estimates.
We average our metrics on all these instances. We study three structures of increasing complexity, as depicted in Figure~\ref{fig:structure_poses_for_eval}, including an essentially  2D arch with little occlusion, for which pose estimation works well, as well as more challenging 3D stacks with several partially or fully hidden objects.

\subsection{Robustness to missing primitives}\label{subsec:eval_structures}
We first verify off-line that our approach is able to cope with incomplete structure information and still find assembly plans that correctly match the target structure. Table \ref{tab:constructions} shows that our algorithm finds assembly plans for every stack in a reasonable number of MCTS rollouts, even for the difficult structure C whose assembly sequences are composed of at least 8 operators.
Results on structure A confirm that assembling a structure whose poses are all known is immediate, with only one rollout required to find the solution. Interestingly, the average reward per rollout decreases as the number of unseen objects grows. As the branching factor of MCTS increases while the number of valid plans remains constant, the planner has to explore more possibilities, resulting in an important slowdown.

\subsection{Ablation study}
\label{subsec:ablations}
\noindent\textbf{Interpenetration removal:}
\label{subsec:eval_collision_removal}
We use here the same setup as in the previous section but
study the effect of different parts of our algorithm. Table \ref{tab:collision_removal} shows that the object penetration removal module is essential for the robustness of our
planner, which may fail up to 35\% of the time without it
in the case of missing primitives. Using this module comes
with little cost, both in terms of time and number of rollouts. It even accelerates the search by eliminating plans with penetrating poses and guiding the search towards higher quality and feasible assemblies in the case of structure C.

\noindent\textbf{Search method and reward:}
\label{subsec:eval_search_reward}
We compare our MCTS approach and \textit{dense} reward definition (Eq. [\ref{eq:reward_formula}]) with both a random search, i.e. replacing the selection with equation (\ref{eq:uct_formula}) by a random choice, and MCTS with a \textit{sparse} reward, equal to one only when all the target poses are matched, i.e. {$r_{sparse}((q_i)_i) = \prod_{i\in\mathcal{V}} 1_{\|q_i^{\mathcal{S}} - q_i^{\mathcal{V}}\|_2 \leq \epsilon}$}.
We also study the impact of guiding the search by favoring, whenever possible, the actions on previously detected objects from $\mathcal{V}$. As we know their target poses, we can readily place them in the assembly with confidence.
For the structure A whose poses are all known, no search is required and all the methods are equivalent.
Table \ref{tab:mcts_random_search} shows that guiding the search
for structures B and C yields a factor of magnitude speedup
in terms of number of rollouts. Except in one case for structure B, MCTS always outperforms random search, quite significantly for structure C. Using a dense reward instead of a sparse one often improves performance, especially for structure C, but
by a relatively small margin.

\subsection{Experiments on a real robot \label{subsec:exp_real}}
\input{figures/other_structures}

\noindent\textbf{Assemblies:}
\label{subsec:eval_robot_real}
We build our structures with a real robot manipulator, where we assemble 10 instances of structures A and B and 15 of structure C (Fig. \ref{fig:structure_poses_for_eval}) and report results in Table \ref{tab:real_robot}.
First, the number of detected objects varies between instances of the same structure: several false positive detections may appear while other objects go undetected. For instance, CosyPose detects an extra object in one instance of structure A, accounting for 10\% of reconstruction failures. In general, detection-related errors represent 10\% to 20\% of failures in our study.
Still, the method remains efficient and succeeds in finding plans in a reasonable number of rollouts when pose estimation detects a sufficient number of objects. Note here that the variance on the number of rollouts is high, due to the changing number of detected objects between instances of a category, which impacts the difficulty of planning.

The reconstruction success rate is 90\% for structure A, 60\% for B and 13\% for C. In every case, structures that are correctly assembled and stable also match the target objects well. 
We have identified 4 sources of failures during the assembly.
First, bad pose estimations on the target images occur in 20\% of the assemblies and prevent planning.
Second, inaccurate robot moves combined with plans placing objects close to the edge of objects underneath lead to falls in respectively 20\% and 13\% of instances of B and C.
Third, 20\% of plans for structure C place objects too close to each other, and our motion planner is unable in these cases to find collision-free paths for placing them and then opening the gripper.
Last, our approximation of objects by boxes leads to failures for 27\% of assemblies of C: our planner generates stable structures given the box models, but these are too large compared to the real objects, causing the structure to collapse during assembly. 

\noindent\textbf{Qualitative results:}
Figure \ref{fig:other_structures} shows examples of some additional structures that our method is able to assemble from an image, a bigger 2D arch as well as more complex 3D stacks with occluded objects. Our method deals with a wide range of assemblies and recovers the pose of missing objects in physically plausible poses that lead to successful assemblies on a real robot.
Our supplementary video includes several examples of assembly runs.


\section{Conclusion}

This work introduces a method to assemble blocks and reproduce structures specified by a single RGB image. 
We find assembly plans from incomplete information about the structure due to unavoidable occlusions by leveraging physical reasoning and show that our method assembles a variety of structures on a real robot. 
While we focus on assembling in a 3D Manhattan space, this is not a limitation of the approach and we believe more complex structures made of diverse shapes can be assembled with our method by using richer STRIPS operators and formulating general stability constraints to handle more complex contacts. Improvements may also come from the use of a better pose estimator, possibly relying on multiple target views, to recover higher quality poses and plan larger assemblies.

\section*{Acknowledgement}

This work was supported in part by the Inria/NYU collaboration, the Louis Vuitton/ENS chair on artificial intelligence and the French government under management of Agence Nationale de la Recherche as part of the "Investissements d’avenir" program, reference ANR19-P3IA0001 (PRAIRIE 3IA Institute). The authors warmly thank Yann Labbé for his help in running CosyPose on the UR5 manipulator and the reviewers for their useful comments.


\bibliographystyle{plain}
\bibliography{biblio}

\end{document}

%% file: figures/teaser.tex
\begin{figure}
    \centering
    \vspace{2mm}
    \includegraphics[width=\linewidth]{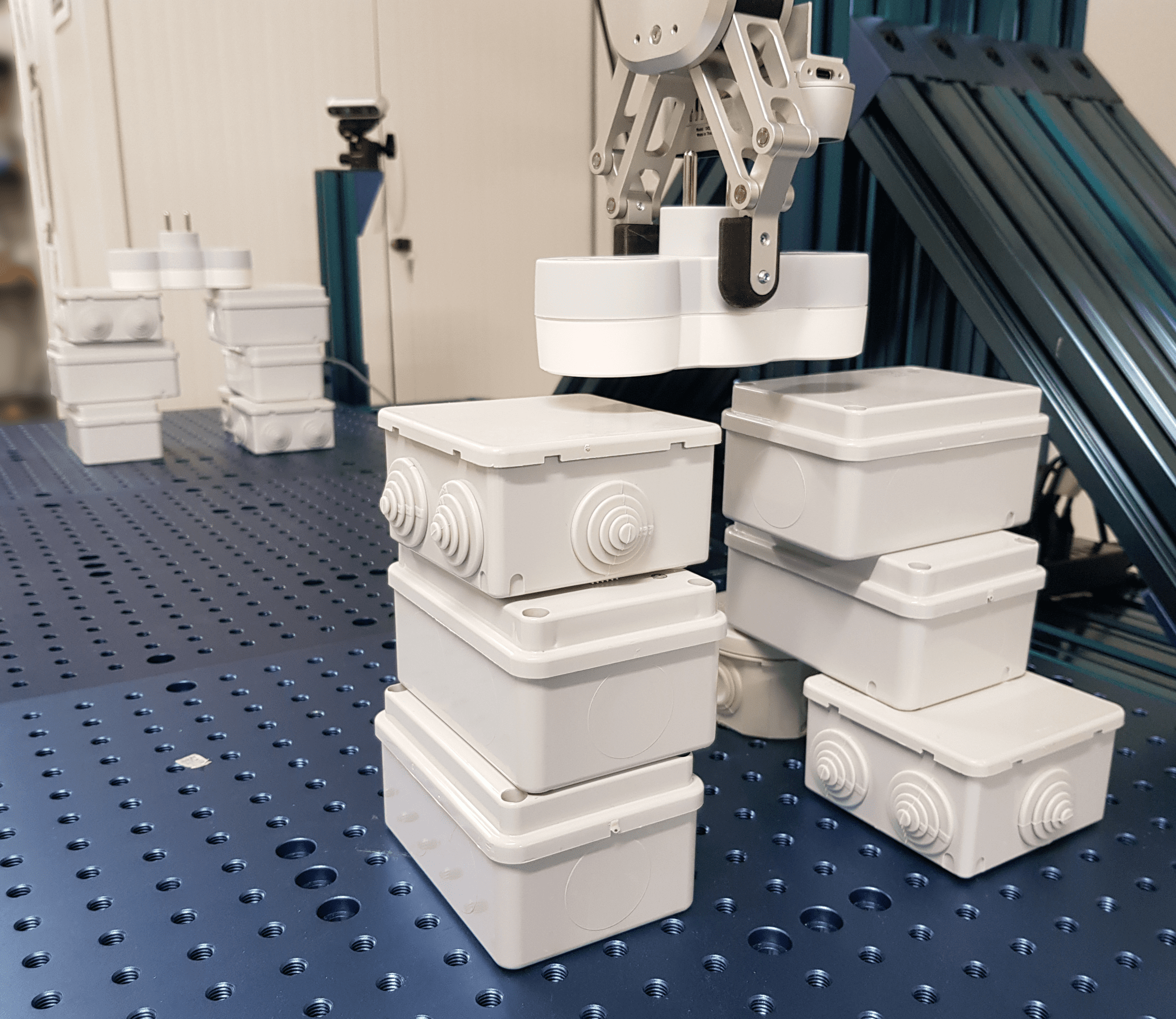}
    \caption{Example result of our approach. A UR5 manipulator assembles a configuration of known primitives (right - foreground). The configuration is specified by a single photograph of the  assembly (camera on the left - background).}
    \vspace*{-3.5mm}
    \label{fig:teaser}
\end{figure}

%% file: figures/overview.tex
\begin{figure*}[thpb]
    \centering
    \includegraphics[width=\textwidth]{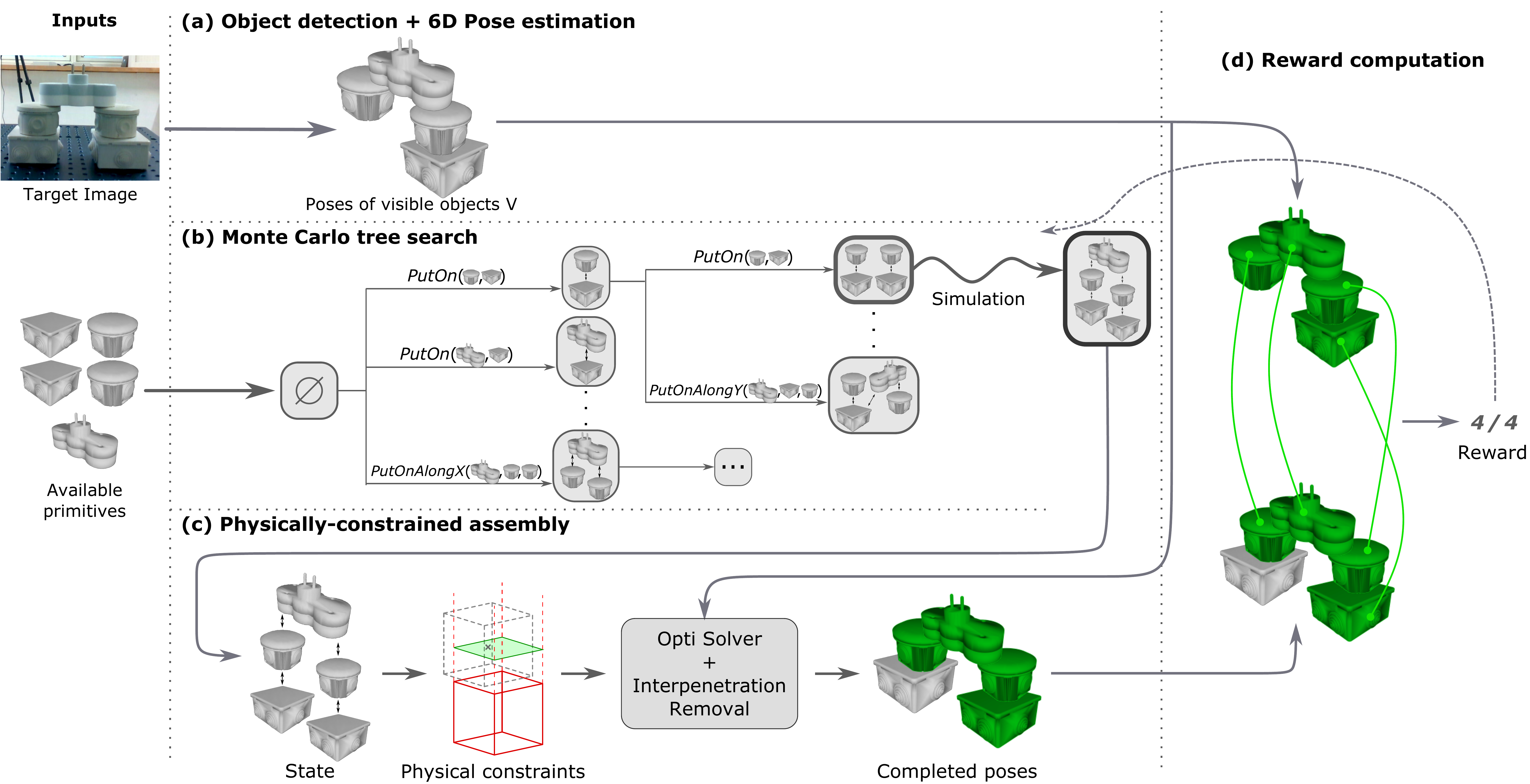}
    \caption{Overview of our method. (a) Given an image of the assembly to build, we extract 6D pose estimations of partially visible objects. (b) An assembly plan of the primitives is built as a sequence of STRIPS operators guided by a Monte Carlo tree search. (c) The precise pose of each primitive is computed using the assembly plan and physical feasibility constraints. (d) The assembled primitives are compared to the poses extracted from the picture in~(a). The difference is used as a reward for MCTS.}
    \vspace*{-3mm}
    \label{fig:overview}
\end{figure*}

%% file: figures/dataset.tex
\begin{figure}
    \centering
    \includegraphics[width=\linewidth]{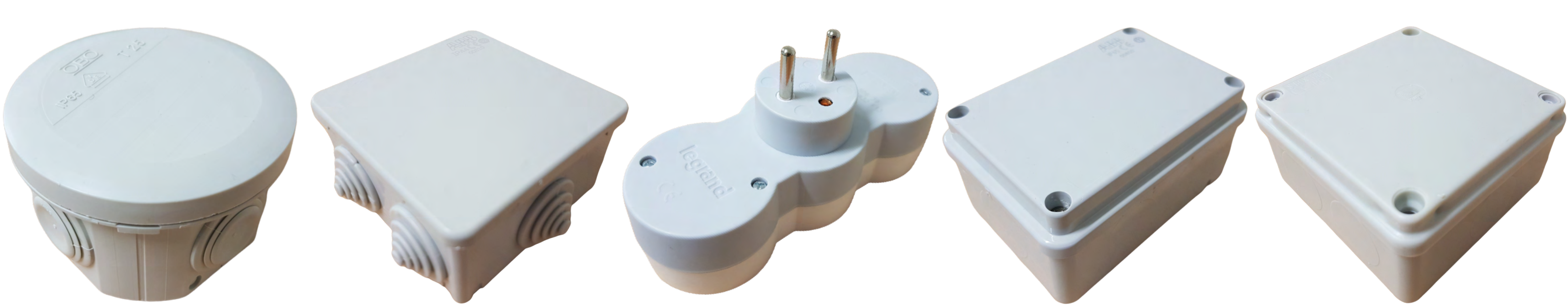}
    \caption{Objects from the T-LESS dataset \cite{hodan2017} used in our constructions.}
    \label{fig:dataset}
    \vspace*{-5.5mm}
\end{figure}

%% file: figures/structures_eval.tex
\begin{figure}
    \centering
    \includegraphics[width=\linewidth]{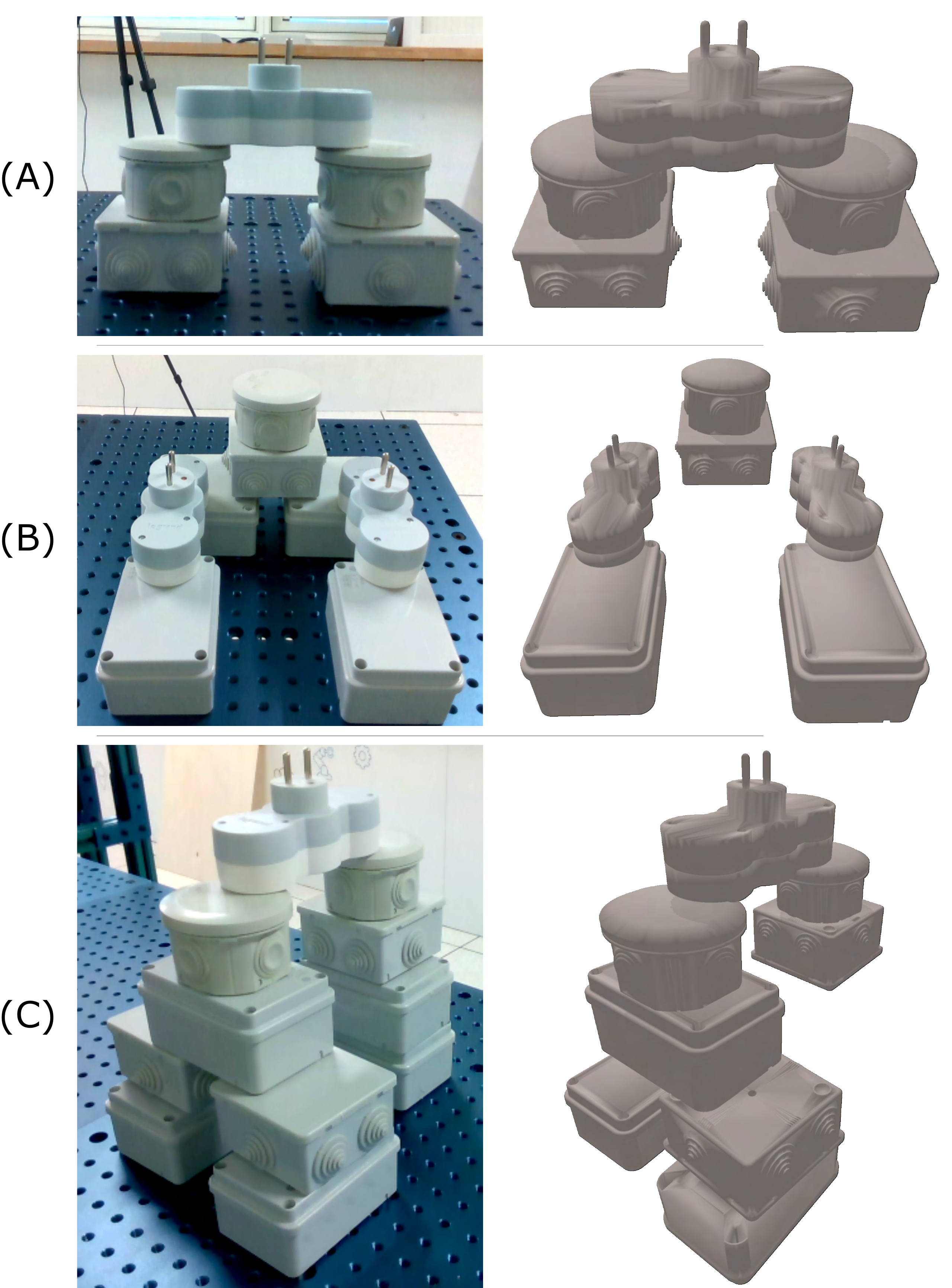}
    \caption{Structures used for our evaluations. Left column: target images; right column: objects and poses seen by the detector and pose estimator. From top to bottom: structures A to C.}
    \label{fig:structure_poses_for_eval}
\end{figure}

%% file: tables/constructions.tex
\begin{table}[ht]
\vspace*{-2mm}
\centering
\caption{Planning computed for various structures on poses retrieved by CosyPose, averaged on 20 random seeds.}
\label{tab:constructions}
\begin{tabular}{lccc}
\toprule
    & \multicolumn{3}{c}{Structures}\\ \cmidrule(lr){2-4}
    & A & B & C \\
\midrule
\begin{tabular}[c]{@{}l@{}}Number of\\ detected objects\end{tabular} & 5 / 5 & 6 / 8 & 8 / 11 \\
\midrule
Success Rate (\%) & 100 & 100 & 100 \\
\midrule
\begin{tabular}[c]{@{}l@{}}Number of\\ MCTS rollouts \end{tabular}
             & 1$\pm$0 & 159$\pm$136 & 882$\pm$686 \\
\midrule
\begin{tabular}[c]{@{}l@{}}Computation\\ time (s) \end{tabular}
             & 0.04$\pm$0.01 & 3.0$\pm$2.5 & 34.0$\pm$26.1 \\
\midrule
\begin{tabular}[c]{@{}l@{}}Average reward\\ per rollout \end{tabular}
             & 1.0$\pm$0.0 & 0.04$\pm$0.07 & 0.011$\pm$0.008 \\
\bottomrule
\end{tabular}
\vspace*{-3.5mm}
\end{table}

%% file: tables/collision_removal.tex
\begin{table}[ht]
\centering
\caption{Effectiveness of our penetration removal module to obtain feasible plans. Metrics are averaged over 20 random seeds.}
\label{tab:collision_removal}
\begin{tabular}{lcccc}
\toprule &
    \multirow{2}{*}{\begin{tabular}[c]{@{}c@{}}Collision\\ Removal\end{tabular}}
    & \multicolumn{3}{c}{Structures}\\ 
    \cmidrule(lr){3-5}
    & \multicolumn{1}{c}{}
    & \multicolumn{1}{c}{A}
    & \multicolumn{1}{c}{B}
    & \multicolumn{1}{c}{C} \\
\midrule
\multirow{2}{*}{\begin{tabular}[c]{@{}l@{}}Success\\ Rate (\%) \end{tabular}}
    & \xmark & 100 & 65  & 70 \\
    & \cmark & 100 & 100 & 100 \\
\midrule
\multirow{2}{*}{\begin{tabular}[c]{@{}l@{}}Number of\\ rollouts\end{tabular}}
    & \xmark & 1$\pm$0 & 124$\pm$89 & 1109$\pm$818 \\
    & \cmark & 1$\pm$0 & 159$\pm$136 & 882$\pm$686 \\
\midrule
\multirow{2}{*}{\begin{tabular}[c]{@{}l@{}}Computation\\ time (s)\end{tabular}}
    & \xmark & 0.05$\pm$0.02 & 1.6$\pm$1.2 & 39.2$\pm$28.8 \\
    & \cmark & 0.05$\pm$0.01 & 3.0$\pm$2.5 & 34.0$\pm$26.1 \\
\bottomrule
\end{tabular}
\vspace*{-2mm}
\end{table}

%% file: tables/mcts_random_search.tex
\begin{table}[ht]
\centering
\vspace{2mm}
\caption{Number of rollouts for Random Search (R.S.) or MCTS with sparse or dense reward and guidance. All the methods have a success rate of 100\% and the metrics are averaged over 20 random seeds.}
\label{tab:mcts_random_search}
\begin{tabular}{lccccc}
\toprule
    \multirow{2}{*}{\begin{tabular}[c]{@{}c@{}}Search\\method\end{tabular}}
    & \multirow{2}{*}{\begin{tabular}[c]{@{}c@{}}Reward\end{tabular}}
    & \multirow{2}{*}{\begin{tabular}[c]{@{}c@{}}Guided\end{tabular}}
    & \multicolumn{3}{c}{Structures} \\ \cmidrule(lr){4-6}
    & & & A & B & C \\
\midrule
    R.S. & --- & \xmark & 1$\pm$0 & 133$\pm$87 & 7100$\pm$6931 \\
    R.S. & --- & \cmark & 1$\pm$0 & 148$\pm$116 & 2077$\pm$2237 \\
    MCTS & Sparse & \xmark & 1$\pm$0 & 99$\pm$89 & 10980$\pm$12686 \\
    MCTS & Sparse & \cmark & 1$\pm$0 & 106$\pm$96 & 1139$\pm$810 \\
    MCTS & Dense & \xmark & 1$\pm$0 & 111$\pm$117& 10346$\pm$7720 \\
    MCTS & Dense & \cmark  & 1$\pm$0 & 159$\pm$136 & 882$\pm$686 \\
\bottomrule
\end{tabular}
\end{table}

%% file: tables/real_robot.tex
\begin{table}[ht]
\centering
\caption{Reconstructions on a UR5 robot manipulator, averaged on 10 instances per structure for A and B and 15 for C.}
\label{tab:real_robot}
\begin{tabular}{lccc}
\toprule & \multicolumn{3}{c}{Structures} \\
    \cmidrule(lr){2-4}
    & A & B & C \\
\midrule
\begin{tabular}[c]{@{}l@{}}Number of\\ visible objects \end{tabular} & 5.1$\pm$0.3 & 6.2$\pm$0.4 & 7.7$\pm$0.7 \\
\midrule
\begin{tabular}[c]{@{}l@{}}Valid initial poses rate (\%) \end{tabular}
             & 90 & 80 & 80 \\
\midrule
\begin{tabular}[c]{@{}l@{}}Number of\\ MCTS rollouts \end{tabular}
             & 1$\pm$0 & 138$\pm$137 & 557$\pm$1050 \\
\midrule
\begin{tabular}[c]{@{}l@{}}Structural stability rate (\%) \end{tabular}
             & 90 & 60 & 13 \\
\midrule
\begin{tabular}[c]{@{}l@{}}Target objects\\matching rate (\%) \end{tabular}
             & 90 & 60 & 13 \\
\bottomrule
\end{tabular}
\vspace*{-2mm}
\end{table}

%% file: figures/other_structures.tex
\begin{figure}
    \centering
    \includegraphics[width=\linewidth]{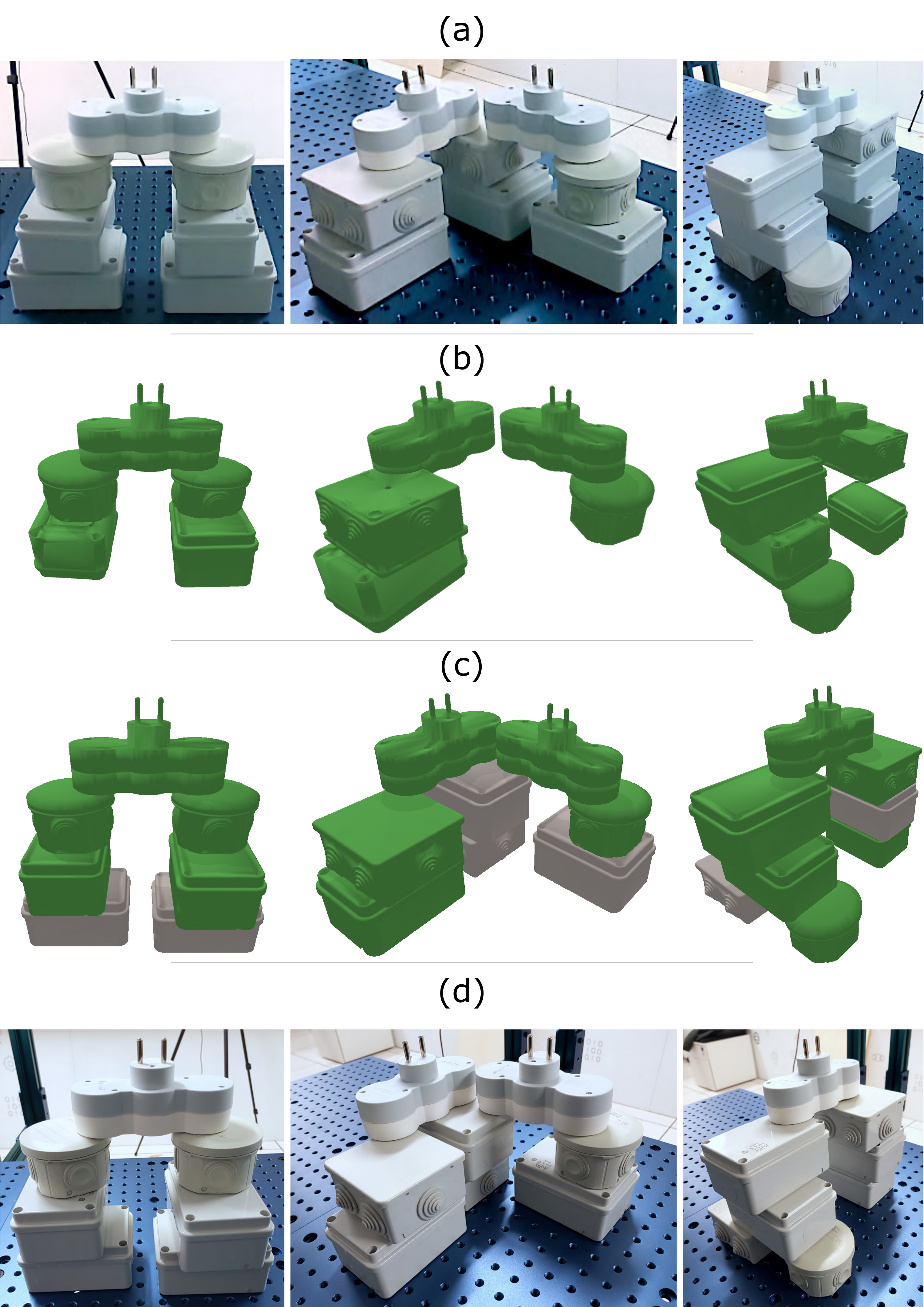}
    \caption{Examples of additional structures rebuilt with our method. (a) Input image of the structure to be assembled,
    (b) identification and 6D pose estimation with CosyPose,
    (c) result of our task planning, where green objects were detected by the pose estimator and grey ones were added by our planning,
    (d) result of the assembly with a real robot.
    }
    \label{fig:other_structures}
    \vspace*{-5mm}
\end{figure}

%% file: root.bbl
\begin{thebibliography}{10}

\bibitem{agrawal2018}
Akshay Agrawal, Robin Verschueren, Steven Diamond, and Stephen Boyd.
\newblock A rewriting system for convex optimization problems.
\newblock {\em Journal of Control and Decision}, 5(1):42--60, 2018.

\bibitem{behnel2011}
Stefan Behnel, Robert Bradshaw, Craig Citro, Lisandro Dalcin, Dag~Sverre
  Seljebotn, and Kurt Smith.
\newblock Cython: The best of both worlds.
\newblock {\em Computing in Science \& Engineering}, 13(2):31--39, 2011.

\bibitem{diamond2016}
Steven Diamond and Stephen Boyd.
\newblock {CVXPY}: {A} {P}ython-embedded modeling language for convex
  optimization.
\newblock {\em Journal of Machine Learning Research}, 17(83):1--5, 2016.

\bibitem{driess2020b}
Danny Driess, Jung{-}Su Ha, and Marc Toussaint.
\newblock Deep visual reasoning: Learning to predict action sequences for task
  and motion planning from an initial scene image.
\newblock In {\em Robotics: Science and Systems}, 2020.

\bibitem{driess2020a}
Danny Driess, Ozgur Oguz, Jung{-}Su Ha, and Marc Toussaint.
\newblock Deep visual heuristics: Learning feasibility of mixed-integer
  programs for manipulation planning.
\newblock In {\em {ICRA}}, pages 9563--9569. {IEEE}, 2020.

\bibitem{fikes1971}
Richard Fikes and Nils~J. Nilsson.
\newblock {STRIPS:} {A} new approach to the application of theorem proving to
  problem solving.
\newblock {\em Artif. Intell.}, 2(3/4):189--208, 1971.

\bibitem{funk2021}
Niklas Funk, Georgia Chalvatzaki, Boris Belousov, and Jan Peters.
\newblock Learn2assemble with structured representations and search for robotic
  architectural construction.
\newblock In {\em CoRL}, volume 164 of {\em Proceedings of Machine Learning
  Research}, pages 1401--1411. {PMLR}, 2021.

\bibitem{garrett2020}
Caelan~Reed Garrett, Tom{\'{a}}s Lozano{-}P{\'{e}}rez, and Leslie~Pack
  Kaelbling.
\newblock Pddlstream: Integrating symbolic planners and blackbox samplers via
  optimistic adaptive planning.
\newblock In {\em {ICAPS}}, pages 440--448. {AAAI} Press, 2020.

\bibitem{pddl}
Malik Ghallab, Craig Knoblock, David Wilkins, Anthony Barrett, Dave
  Christianson, Marc Friedman, Chung Kwok, Keith Golden, Scott Penberthy, David
  Smith, Ying Sun, and Daniel Weld.
\newblock Pddl - the planning domain definition language.
\newblock 08 1998.

\bibitem{he2017}
Kaiming He, Georgia Gkioxari, Piotr Doll{\'{a}}r, and Ross~B. Girshick.
\newblock Mask {R-CNN}.
\newblock In {\em {ICCV}}, pages 2980--2988. {IEEE} Computer Society, 2017.

\bibitem{he2020}
Yisheng He, Wei Sun, Haibin Huang, Jianran Liu, Haoqiang Fan, and Jian Sun.
\newblock {PVN3D:} {A} deep point-wise 3d keypoints voting network for 6dof
  pose estimation.
\newblock In {\em {CVPR}}, pages 11629--11638. Computer Vision Foundation /
  {IEEE}, 2020.

\bibitem{hodan2017}
Tomas Hodan, Pavel Haluza, Step{\'{a}}n Obdrz{\'{a}}lek, Jiri Matas, Manolis
  I.~A. Lourakis, and Xenophon Zabulis.
\newblock {T-LESS:} an {RGB-D} dataset for 6d pose estimation of texture-less
  objects.
\newblock In {\em {WACV}}, pages 880--888. {IEEE} Computer Society, 2017.

\bibitem{hodan2018}
Tomas Hodan, Frank Michel, Eric Brachmann, Wadim Kehl, Anders~Glent Buch, Dirk
  Kraft, Bertram Drost, Joel Vidal, Stephan Ihrke, Xenophon Zabulis, Caner
  Sahin, Fabian Manhardt, Federico Tombari, Tae{-}Kyun Kim, Jiri Matas, and
  Carsten Rother.
\newblock {BOP:} benchmark for 6d object pose estimation.
\newblock In {\em {ECCV} {(10)}}, volume 11214 of {\em Lecture Notes in
  Computer Science}, pages 19--35. Springer, 2018.

\bibitem{ikeuchi1994}
Katsushi Ikeuchi and Takashi Suehiro.
\newblock Toward an assembly plan from observation. i. task recognition with
  polyhedral objects.
\newblock {\em {IEEE} Trans. Robotics Autom.}, 10(3):368--385, 1994.

\bibitem{kuffner2000}
James J.~Kuffner Jr. and Steven~M. LaValle.
\newblock Rrt-connect: An efficient approach to single-query path planning.
\newblock In {\em {ICRA}}, pages 995--1001. {IEEE}, 2000.

\bibitem{labbe2020b}
Yann Labb{\'{e}}, Justin Carpentier, Mathieu Aubry, and Josef Sivic.
\newblock Cosypose: Consistent multi-view multi-object 6d pose estimation.
\newblock In {\em {ECCV} {(17)}}, volume 12362 of {\em Lecture Notes in
  Computer Science}, pages 574--591. Springer, 2020.

\bibitem{labbe2020a}
Yann Labb{\'{e}}, Sergey Zagoruyko, Igor Kalevatykh, Ivan Laptev, Justin
  Carpentier, Mathieu Aubry, and Josef Sivic.
\newblock Monte-carlo tree search for efficient visually guided rearrangement
  planning.
\newblock {\em {IEEE} Robotics Autom. Lett.}, 5(2):3715--3722, 2020.

\bibitem{li2018}
Yi~Li, Gu~Wang, Xiangyang Ji, Yu~Xiang, and Dieter Fox.
\newblock Deepim: Deep iterative matching for 6d pose estimation.
\newblock In {\em {ECCV} {(6)}}, volume 11210 of {\em Lecture Notes in Computer
  Science}, pages 695--711. Springer, 2018.

\bibitem{lin2021}
Yixin Lin, Austin~S. Wang, Eric Undersander, and Akshara Rai.
\newblock Efficient and interpretable robot manipulation with graph neural
  networks.
\newblock {\em {IEEE} Robotics Autom. Lett.}, 7(2):2740--2747, 2022.

\bibitem{lozano2014}
Tom{\'{a}}s Lozano{-}P{\'{e}}rez and Leslie~Pack Kaelbling.
\newblock A constraint-based method for solving sequential manipulation
  planning problems.
\newblock In {\em {IROS}}, pages 3684--3691. {IEEE}, 2014.

\bibitem{migimatsu2019}
Toki Migimatsu and Jeannette Bohg.
\newblock Object-centric task and motion planning in dynamic environments.
\newblock {\em {IEEE} Robotics Autom. Lett.}, 5(2):844--851, 2020.

\bibitem{pashevich2020}
Alexander Pashevich, Igor Kalevatykh, Ivan Laptev, and Cordelia Schmid.
\newblock Learning visual policies for building 3d shape categories.
\newblock In {\em {IROS}}, pages 8073--8080. {IEEE}, 2020.

\bibitem{paxton17}
Chris Paxton, Vasumathi Raman, Gregory~D. Hager, and Marin Kobilarov.
\newblock Combining neural networks and tree search for task and motion
  planning in challenging environments.
\newblock In {\em {IROS}}, pages 6059--6066. {IEEE}, 2017.

\bibitem{Pednault1987FORMULATINGMD}
Edwin P.~D. Pednault.
\newblock Formulating multiagent, dynamic-world problems in the classical
  planning framework.
\newblock 1987.

\bibitem{rad17}
Mahdi Rad and Vincent Lepetit.
\newblock {BB8:} {A} scalable, accurate, robust to partial occlusion method for
  predicting the 3d poses of challenging objects without using depth.
\newblock In {\em {ICCV}}, pages 3848--3856. {IEEE} Computer Society, 2017.

\bibitem{rothganger2006}
Fred Rothganger, Svetlana Lazebnik, Cordelia Schmid, and Jean Ponce.
\newblock 3d object modeling and recognition using local affine-invariant image
  descriptors and multi-view spatial constraints.
\newblock {\em Int. J. Comput. Vis.}, 66(3):231--259, 2006.

\bibitem{silver2016}
David Silver, Aja Huang, Chris~J. Maddison, Arthur Guez, Laurent Sifre, George
  van~den Driessche, Julian Schrittwieser, Ioannis Antonoglou, Vedavyas
  Panneershelvam, Marc Lanctot, Sander Dieleman, Dominik Grewe, John Nham, Nal
  Kalchbrenner, Ilya Sutskever, Timothy~P. Lillicrap, Madeleine Leach, Koray
  Kavukcuoglu, Thore Graepel, and Demis Hassabis.
\newblock Mastering the game of go with deep neural networks and tree search.
\newblock {\em Nat.}, 529(7587):484--489, 2016.

\bibitem{silver2017}
David Silver, Julian Schrittwieser, Karen Simonyan, Ioannis Antonoglou, Aja
  Huang, Arthur Guez, Thomas Hubert, Lucas Baker, Matthew Lai, Adrian Bolton,
  Yutian Chen, Timothy~P. Lillicrap, Fan Hui, Laurent Sifre, George van~den
  Driessche, Thore Graepel, and Demis Hassabis.
\newblock Mastering the game of go without human knowledge.
\newblock {\em Nat.}, 550(7676):354--359, 2017.

\bibitem{stevsic2020}
Stefan Stevsic, Sammy Christen, and Otmar Hilliges.
\newblock Learning to assemble: Estimating 6d poses for robotic object-object
  manipulation.
\newblock {\em {IEEE} Robotics Autom. Lett.}, 5(2):1159--1166, 2020.

\bibitem{toussaint2015}
Marc Toussaint.
\newblock Logic-geometric programming: An optimization-based approach to
  combined task and motion planning.
\newblock In {\em {IJCAI}}, pages 1930--1936. {AAAI} Press, 2015.

\bibitem{toussaint17}
Marc Toussaint and Manuel Lopes.
\newblock Multi-bound tree search for logic-geometric programming in
  cooperative manipulation domains.
\newblock In {\em {ICRA}}, pages 4044--4051. {IEEE}, 2017.

\bibitem{wang2019}
Chen Wang, Danfei Xu, Yuke Zhu, Roberto~Martin Martin, Cewu Lu, Li~Fei{-}Fei,
  and Silvio Savarese.
\newblock Densefusion: 6d object pose estimation by iterative dense fusion.
\newblock In {\em {CVPR}}, pages 3343--3352. Computer Vision Foundation /
  {IEEE}, 2019.

\bibitem{winston1970}
Patrick~H. Winston.
\newblock Copy demo.
\newblock \url{https://people.csail.mit.edu/bkph/phw\_copy\_demo.shtml}.
\newblock Accessed: 2022-03-01.

\bibitem{xiang2018}
Yu~Xiang, Tanner Schmidt, Venkatraman Narayanan, and Dieter Fox.
\newblock Posecnn: {A} convolutional neural network for 6d object pose
  estimation in cluttered scenes.
\newblock In {\em Robotics: Science and Systems}, 2018.

\end{thebibliography}
